\documentclass[10pt,twocolumn,letterpaper]{article}

\usepackage{cvpr}
\usepackage{times}
\usepackage{epsfig}
\usepackage{graphicx}
\usepackage{amsmath}
\usepackage{amssymb}

\usepackage[breaklinks=true,bookmarks=false]{hyperref}

\cvprfinalcopy % *** Uncomment this line for the final submission

\def\cvprPaperID{****} % *** Enter the CVPR Paper ID here

\begin{document}

%%%%%%%%% TITLE
\title{Pseudo-labels for Supervised Learning on Dynamic Vision Sensor Data, Applied to Object Detection under Ego-motion}

\author{Nicholas F. Y. Chen\\
DSO National Laboratories\\
12 Science Park Drive, Singapore (118225)\\
{\tt\small cfangyew@dso.org.sg}
%\and
}

\maketitle
%\thispagestyle{empty}

%%%%%%%%% ABSTRACT
\begin{abstract}
In recent years, dynamic vision sensors (DVS), also known as event-based cameras or neuromorphic sensors, have seen increased use due to various advantages over conventional frame-based cameras. Using principles inspired by the retina, its high temporal resolution overcomes motion blurring, its high dynamic range overcomes extreme illumination conditions and its low power consumption makes it ideal for embedded systems on platforms such as drones and self-driving cars. However, event-based data sets are scarce and labels are even rarer for tasks such as object detection. We transferred discriminative knowledge from a state-of-the-art frame-based convolutional neural network (CNN) to the event-based modality via intermediate pseudo-labels, which are used as targets for supervised learning. We show, for the first time, event-based car detection under ego-motion in a real environment at 100 frames per second with a test average precision of 40.3\% relative to our annotated ground truth. The event-based car detector handles motion blur and poor illumination conditions despite not explicitly trained to do so, and even complements frame-based CNN detectors, suggesting that it has learnt generalized visual representations.
\end{abstract}

%%%%%%%%% BODY TEXT
\section{Introduction}

\begin{figure*}[t]
\begin{center}
 \includegraphics[width=1.0\linewidth]{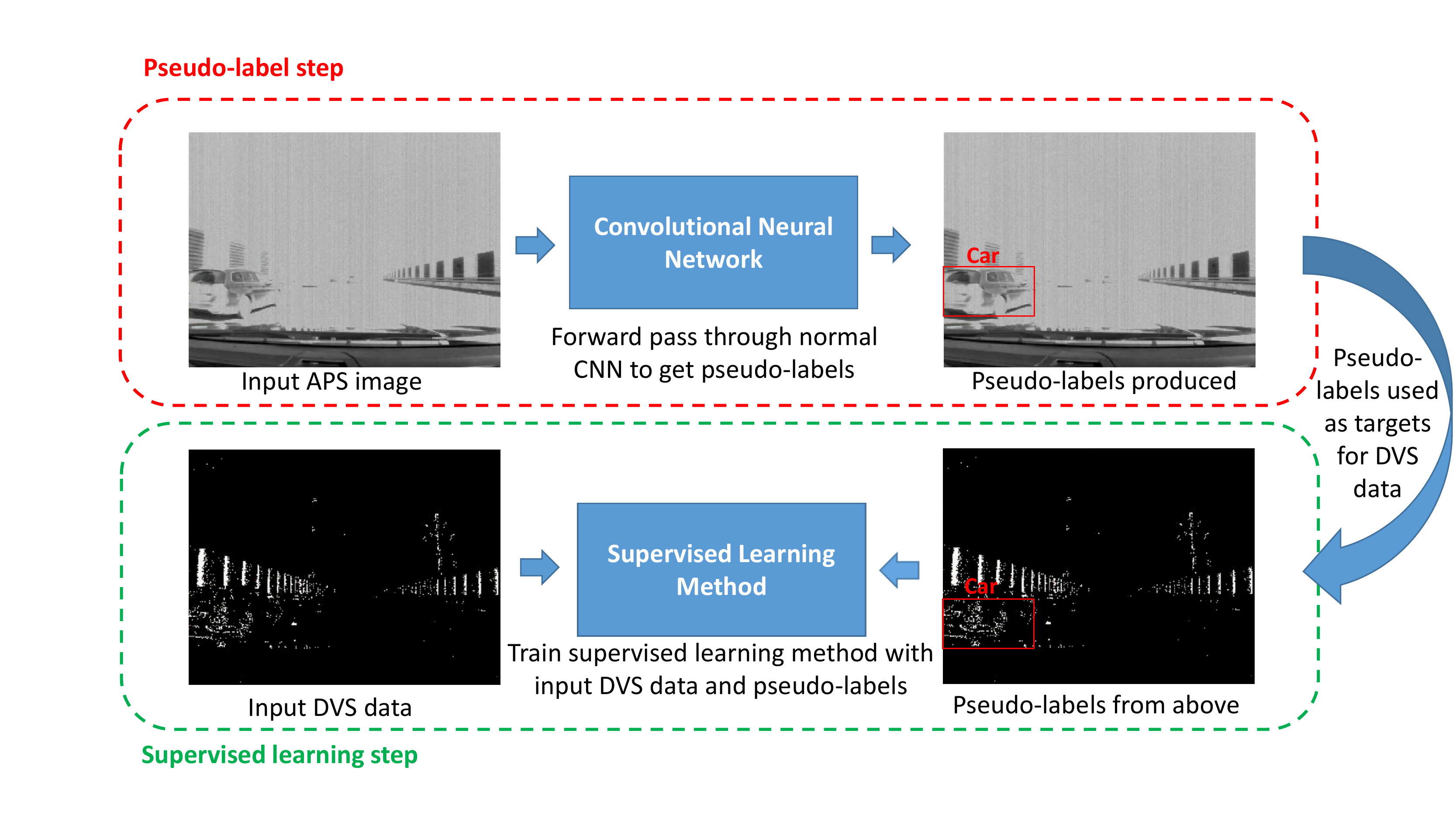}
\end{center}
   \caption{Schematic of the proposed pseudo-labeling and supervised learning method. Top frame: Image from the APS sensor is passed through a CNN to get the pseudo-labels. Bottom frame: Since the dynamic vision sensor is synchronized with the APS (grayscale) camera, the pseudo-labels from the previous step is treated as ground truth to train a supervised learning method which takes dynamic vision sensor data as inputs.}
\label{fig:pseudolabel}
\end{figure*}

Dynamic vision sensors (DVS), also known as event-based cameras or neuromorphic sensors~\cite{ddd2,ddd3}, are a class of biologically-inspired sensors which capture data in an asynchronous manner. When a pixel detects a change in luminance above a certain threshold in log scale, the device emits an output (hence called an `event') containing the pixel location, time and polarity (+1 or -1, corresponding to an increase and decrease in luminance respectively). Such sensors have a temporal resolution on the order of milliseconds or less, making the device suitable for high speed recognition, tracking and collision avoidance. Other advantages of dynamic vision sensors include a high dynamic range and power efficiency, making it ideal for outdoor usage on embedded systems in robotics. 

Frame-based labeled data sets are widely available, contributing to the tremendous advancements in frame-based computer vision in recent years. However, event-based computer vision is still in the process of maturing, and current event-based data sets are quite limited, especially in the case of object detection. Event-based data sets have been released for robotics applications such as simultaneous localization and mapping (SLAM), visual navigation, pose estimation and optical flow estimation~\cite{Barranco2016,Mueggler2017,opticalFlow,EBSLAM}, and they comprise of mostly indoor scenes such as objects on a table top, boxes in a room, posters and shapes, and occasional outdoor scenes. For object recognition and detection, some data sets were created by placing a dynamic vision sensor in front of a monitor and recording existing frame-based data sets~\cite{NMNIST, Hu2016}. Moeys \etal~\cite{Moeys2016} recorded scenes of a predator robot chasing a prey robot in a controlled lab environment with some background objects, and includes ground truth of the prey robot position.

In the long run, dynamic vision sensors will be integrated in platforms such as drones and autonomous vehicles which work in complex, outdoor environments. The DAVIS Driving Dataset 2017 (DDD17) ~\cite{ddd17} is the only data set as of writing which captures such environments, and is the largest event-based data set to date, with over 400 GB and 12 hours worth of driving data spread across over 40 scenes at a resolution of 346 $\times$ 260 pixels. These scenes are varied over the times of the day (day, evening, night), weather (dry, rainy, wet) and location (campus, city, town, freeway, highway), and includes vehicle details like velocity, steering wheel angle and accelerator pedal position. The DAVIS is a camera model which contains a dynamic vision sensor synchronized with a grayscale frame-based camera (also known as the active pixel sensor, or APS).

High speed object detection under ego-motion from dynamic vision sensor data serves a few purposes. First, dynamic vision sensors overcome problems which ordinary frame-based cameras typically encounter. At high speeds, frame-based cameras suffer from motion blur and collision avoidance is limited, placing a speed limit on the platform which the camera is mounted on. In extreme illumination conditions, frame-based cameras have difficulty capturing features of objects. Since dynamic vision sensors output changes in luminance, the data is a sparse representation which can be processed faster, compared to the output of frame-based cameras which contains (potentially redundant) background information. Also, detections from dynamic vision sensor data can be used to complement detections from frame-based cameras, as we will show from our experiments. Last, detection under ego-motion is required because dynamic vision sensors mounted on platforms will inevitably have ego-motion, and the output of the sensors will include some background information as a result, creating distractions which the detection algorithm must overcome.

Like most objects in event-based data sets however, objects in the DDD17 are not labeled. In this paper, we take advantage of the mature state of frame-based detection by using a state-of-the-art CNN to perform car detection on the grayscale (APS) images of the DDD17. These detections, hence termed `pseudo-labels', are shown to be effective when used as targets for a separate (fast) CNN when training on dynamic vision sensor data in the form of binned frames. A schematic of this method can be found in Figure \ref{fig:pseudolabel}.  \\

\noindent\textbf{Contributions}

\begin{enumerate}
\item We trained a CNN on pseudo-labels to detect cars from dynamic vision sensor data, with a test average precision of 40.3\% relative to annotated ground truth. This is the first time that high-speed (100 FPS) object detection is done on dynamic vision sensor data under ego-motion in a real environment, whereas previous works have only focused on recognizing/detecting simple objects in a controlled environment or detecting objects without camera ego-motion.
\item We show that a CNN trained on pseudo-labels can detect cars despite motion blur or poor lighting, even though pseudo-labels were not generated for these scenarios. This CNN even complements the original frame-based CNN that was used to generate the pseudo-labels, suggesting that our trained CNN learnt generalized visual representations of cars.
\end{enumerate}

\subsection{Related work}

\noindent\textbf{Pseudo-labels \& cross modal distillation}\quad Pseudo-labeling was introduced by Lee~\cite{lee2013} for semi-supervised learning on frame-based data, where during each weight update, the unlabeled data picks up the class which has the maximum predicted probability and treats it as the ground truth. Chen \etal~\cite{dcs} proposed a method to incrementally select reliable unlabeled data to give pseudo-labels to. Saito \etal~\cite{saito2017} proposed using three classifiers that regulate each other, to achieve domain adaptation from pseudo-labels. Pathak \etal~\cite{pathak2016} used automatically generated masks (pseudo-labels in their context) from unsupervised motion segmentation on videos, and then trained a CNN to predict these masks from static images. The trained CNN learnt feature representations and was able to perform image classification, semantic segmentation and object detection. 

For data sets with paired modalities (e.g. RGB-D data contains RGB data of a scene synchronized with depth data of the same scene), cross modal distillation~\cite{crossmodal} is a scheme that transfers knowledge from one modality, which has a lot of labels, to another modality, which has very few labels. In ~\cite{crossmodal}, mid-level representations of a CNN trained on RGB images were used to supervise training for another CNN to perform object detection and segmentation on depth images. In~\cite{soundnet}, the visual modality of videos was used to generate pseudo-labels from CNNs and used to train a separate 1-D CNN to classify scenes from sound inputs. Our work is inspired by these cross modal methods, and we leverage on the fact that the DDD17 is a large data set with synchronized DVS and APS modalities.

\smallskip

\noindent\textbf{Event-based object detection}\quad Object detection on dynamic vision sensor data is relatively new since labeled event-based data sets are scarce. Liu \etal~\cite{Liu2016} performed object detection on the predator-prey data set~\cite{Moeys2016}. They used dynamic vision sensor data as an attention mechanism for a frame-based CNN, and compared it to using a CNN to perform detection on the entire grayscale image. Including particle filter for both methods to aid tracking, the former method is 70X faster than the latter, with an accuracy of 90\%. Li \etal~\cite{ATP} proposed a method which adaptively pools feature maps from successive frames (generated by binning dynamic vision sensor data over time) to create motion invariant features for object detection. They demonstrated hand detection on a private data set, with performance scores averaging from 61.3\% to 76.0\% depending on the variant of the method used. Hinz \etal~\cite{hinz2017} demonstrated a tracking-by-clustering system which detects and tracks vehicles on a highway bridge. Both~\cite{ATP} and~\cite{hinz2017} did not benchmark their methods on dynamic vision sensor data under camera ego-motion.

\section{Generating pseudo-labels for dynamic vision sensor data}

\begin{table}
\begin{center}
%\begin{tabular}{|l|c|}
\begin{tabular}{llll}
\hline
Recording  & Scene & Condition & Type\\
\hline\hline
rec1487337800 & campus             & day        & test (1)  \\
rec1487424147 & mostly fwy     & day        & train                   \\
rec1487593224 & hwy            & day        & train                   \\
rec1487597945 & cty               & evening    & train                   \\
rec1487608147 & fwy            & evening    & test (2)  \\
rec1487609463 & fwy            & evening    & train                   \\
rec1487778564 & campus             & day        & val                  \\
rec1487779465 & cty+hwy   & day        & train                    \\
rec1487781509 & campus             & evening    & train                   \\
rec1487782014 & cty+hwy   & evening    & test (3)  \\
rec1487839456 & cty               & day, sunny & train                   \\
rec1487842276 & cty               & day, sunny & train                   \\
rec1487844247 & cty               & day, sunny & train                   \\
rec1487846842 & towns+hwys & day, sunny & val                    \\
rec1487849151 & town               & day, sunny & train                   \\
rec1487849663 & towns+hwys & day, sunny & train                   \\
rec1487856408 & town               & day, sunny & test (4)  \\
rec1487857941 & town               & day, sunny & train                   \\
rec1487858093 & cty               & day, sunny & train                   \\
rec1487860613 & cty               & day, sunny & train                   \\
rec1487864316 & cty+fwy   & evening    & val 	\\
\hline
\end{tabular}
\end{center}
\caption{Details of the recordings used from DDD17 in our experiments. Keys: cty=city, fwy=freeway, hwy=highway. Some scenes from the DDD17 were not used for reasons such as APS frames being too dark (especially at night) or too bright, low DVS sensitivity, errors extracting the data and scenes being too short.}
\label{table:rec}
\end{table}

We overcome the lack of labeled dynamic vision sensor data by using cross modal distillation with pseudo-labels on the DDD17 data set (see Figure~\ref{fig:pseudolabel} for a brief outline). Since the DAVIS sensor has a frame-based camera (APS) synchronized with a dynamic vision sensor, the ground truth in one camera is the same as the ground truth in the other camera. We make use of this correspondence--The grayscale (APS) images are fed into a state-of-the-art CNN which generates outputs (pseudo-labels). These pseudo-labels with confidence above a threshold are treated as ground truth and used to train a supervised learning method, which takes the dynamic vision sensor data as inputs. Though the pseudo-labels are noisy, Pathak \etal~\cite{pathak2016} argues that in the absence of systematic errors, such ‘noise’ are perturbations around the ground truth, and since supervised learning methods like neural networks have a finite capacity, it cannot learn the noise perfectly and it might learn something closer to the ground truth. In the context of our experiments (car detection), the pseudo-labels are bounding boxes while the supervised learning method is also a CNN. Pseudo-labeling is not limited to object detection--it should work for other computer vision tasks like image segmentation, image recognition and activity recognition.

\smallskip

\noindent\textbf{Implementation Details}\quad We chose the Recurrent Rolling Convolution (RRC)~\cite{RRC} CNN as the object detection CNN for APS images because as of writing, it is  the best-performing model on the KITTI Object Detection Evaluation benchmark~\cite{KITTI}. Two versions of the RRC are used: The original model trained on the KITTI data set (which is in RGB), and another model which is fine-tuned over 1000 iterations on a grayscale-converted KITTI data set. This is to investigate the impact of pseudo-labels with different performance. As the RRC takes in images of a different aspect ratio than the APS images, we scaled the APS images to the largest possible size while preserving the aspect ratio, and padded the remainder of the image with zeroes. By keeping predictions that have at least a 0.5 confidence score, we produced about 330k and 400k pseudo-labeled images from the original and fine-tuned RRC respectively for various day and evening scenes (the RRC might not produce accurate detections for the night scenes). The scenes are split into train/val/test sets in the ratio 71/15/14 by their recording length, with each set covering a variety of conditions and scenes. Details of the recordings used from the DDD17 can be found in table \ref{table:rec}. We focused only on detecting cars, but this method can easily be extended to other classes such as pedestrians and cyclists.

\section{Supervised learning with pseudo-labels}

\noindent\textbf{Implementation Details}\quad We adopt a frame-based approach to the dynamic vision sensor data for object detection, because frame-based object detection is mature. The dynamic vision sensor data are converted to images by binning the dynamic vision sensor outputs in 10 ms intervals, and each pixel takes the value

\begin{equation}
\sigma(x) = 255 * \frac{1}{1+e^{-x/2}},
\end{equation}

\noindent where $x$ is the sum of the polarities of the events in the 10 ms interval. We refer to this as the \textit{sigmoid representation} of the dynamic vision sensor data. 10 ms was chosen because we aim to achieve detection at 100 frames per second (FPS), about an order of magnitude above most state-of-the-art CNNs.

We used the tiny YOLO CNN~\cite{yolo1,yolo2} as it is one of the few CNNs that can run above 100 FPS with a decent performance (57.1 mean average precision on the VOC 2007+2012 benchmark). We started with this CNN pre-trained on the VOC 2007+2012 benchmark and fine-tuned it using the pseudo-labels generated, in steps of 10k iterations, up to 150k iterations (including the 20k iterations from pre-training). As we want to show that the object detection CNN performs well as a result of the effectiveness of pseudo-labels rather than the result of optimizing hyper-parameters, we only changed the subdivisions from 8 to 4 and batch size from 64 to 128, and kept the other settings as provided in~\cite{yolo1,yolo2}.

\subsection{Quantitative results}

The scenario that we are tackling (high-speed object detection in a real environment from dynamic vision sensor data under camera ego-motion) is the first of its kind, so there are no other state-of-the-art algorithms for comparison. As such, we hope that this work serves as a benchmark for future methods tackling the same scenario. 

Since there is no ground-truth data for the objects in DDD17, we measure performance relative to the RRC pseudo-labels during the model validation step. The model with the highest average precision on the validation set will then be evaluated on the test set. We use an intersection-over-union (IoU) threshold 0.5 for this step.

\begin{table}
\begin{center}
\begin{tabular}{cccc}
\hline
Modality & Arch. & AP@0.5 & AP@0.7\\
\hline\hline
APS & RRC & 44.1\% & 39.6\% \\
DVS & t.YOLO & 36.9\% & 18.3\% \\
APS+DVS & RRC+t.YOLO & 55.6\% & 39.9\% \\
\hline
APS & RRC(ft) & 53.7\% & 47.2\% \\
DVS & t.YOLO(ft) & $\mathbf{40.3\%}$ & $\mathbf{19.9\% }$\\
APS+DVS & RRC+t.YOLO(ft) & 62.2\% & 47.7\% \\
\hline
\end{tabular}
\end{center}
\caption{Evaluation results of our experiments, at IoU thresholds of 0.5 and 0.7. Keys: Arch.=Architecture, ft=fine-tune, AP=average precision, t.YOLO(ft)= tiny YOLO model trained on pseudo-labels produced by RRC (fine-tuned). We can see that fine-tuning the RRC to grayscale data enables it to generate better pseudo-labels and hence improve performance on the DVS-only detector. Also, the combination of the APS (grayscale) and DVS cameras help achieve a higher performance than either modality alone.}
\label{table:summary}
\end{table}

\smallskip

\noindent\textbf{Evaluation against ground truth}\quad We randomly selected 1000 frames from the test set for manual annotation, and all performance figures reported henceforth are obtained by evaluation on this subset. Similar to the KITTI object detection benchmark, we only consider objects that have a minimum height of 25 pixels. A summary of the results can be found in table \ref{table:summary}. The test average precision of the DVS-only detector is 36.9\% and 40.3\% for pseudo-labels generated by the RRC (original) and the RRC (fine-tuned) respectively, at an IoU threshold of 0.5. As a comparison, the tiny YOLO architecture achieves a mean average precision of 57.1\% when trained on real labels (VOC 2007+2012 benchmark). We see that the model trained on RRC (fine-tuned) pseudo-labels is superior to the model trained on RRC (original) pseudo-labels, which is in line with our expectations because a fine-tuned model will produce more accurate pseudo-labels. Furthermore, the weaker performance of the RRC (original) caused it to produce less pseudo-labels on the training scenes, which could also be a factor in the DVS-only model's weaker performance relative to its RRC (fine-tuned) counterpart. Note that the RRC's performance on our test set is much lower than that reported for the KITTI data set (87.4\% for the hard setting, IoU threshold at 0.7) because the data set we are using has a lower resolution (346 $\times$ 260).

\smallskip

\noindent\textbf{Complementing DVS and grayscale detections}\quad We evaluated if the combination of DVS and grayscale detections can improve the overall performance, listed as APS+DVS in table \ref{table:summary}. We combined the detections of the DVS-only detector and the RRC, and applied non-maximum suppression with an IoU threshold of 0.4 to remove duplicates. At a detection IoU threshold of 0.5, such a combination yielded an average precision of 62.2\% on our annotated ground truth data set, roughly a 16\% increase over using only the RRC. This is despite the fact that the DVS-only detector is trained only on knowledge generated by the RRC,  showing that the DVS-only detector has learnt generalized representations of cars. A similar effect was observed in~\cite{crossmodal} for the RGB and depth modalities.

Given the current state of hardware, the RRC is not a real-time detector and the specific combination of the detections mentioned above is not practical yet. However, we hope that this experiment will inspire future work on using detections from the DVS to complement detections from the APS. 

We notice that at an IoU threshold of 0.7, the benefit from combining the detectors is marginal. This is due to the fact that the RRC architecture is specifically designed to work well at high IoU thresholds, whereas the tiny YOLO architecture is designed assuming that it will be evaluated at an IoU threshold of 0.5.

\smallskip

\noindent\textbf{Comparing DVS and grayscale detections}\quad We measured the correct detections made by the detectors (regardless of the confidence score) as a fraction of the total number of ground truth objects in table \ref{table:frac}. We also take a look at the union and intersection of these detections. At 0.5 IoU threshold, the DVS-only detector picked out 60.1\% of the objects while the RRC picked out 64.2\% of the objects. 10.6\% of the objects were detected by the DVS-only detector but not by the RRC, reinforcing the fact that the DVS-only detector learnt general representations of cars, though it was trained on the knowledge from the RRC. We notice that fine-tuning the RRC did not change the fraction by much for the DVS and APS$\cup$DVS modalities though  it improved the average precision in table \ref{table:summary}--This might be due to the fine-tuning process increasing the confidence of correct detections rather than the number of correct detections made by the DVS-only detector. 

\smallskip

\begin{table}
\begin{center}
\begin{tabular}{cccc}
\hline
Modality & Arch. & Frac.@0.5 & Frac.@0.7\\
\hline\hline
APS & RRC & 55.8\% & 49.5\% \\
DVS & t.YOLO & 61.4\% & 33.0\% \\
APS$\cap$DVS & RRC+t.YOLO & 41.7\% & 25.2\% \\
APS$\cup$DVS & RRC+t.YOLO & 75.4\% & 57.3\% \\
\hline
APS & RRC(ft) & 64.2\% & 55.1\% \\
DVS & t.YOLO(ft) & 60.1\% & 32.4\% \\
APS$\cap$DVS & RRC+t.YOLO(ft) & 49.5\% & 27.1\% \\
APS$\cup$DVS & RRC+t.YOLO(ft) & 74.8\% & 60.4\% \\
\hline
\end{tabular}
\end{center}
\caption{Correct detections made by the detectors as a fraction of all the actual objects. Keys: Arch.=Architecture, ft=fine-tune, Frac.=fraction, t.YOLO(ft)= tiny YOLO model trained on pseudo-labels produced by RRC (fine-tuned). Notice that the DVS detected some objects which are not detected by the RRC.}
\label{table:frac}
\end{table}

\subsection{Qualitative results}

\begin{table}
\begin{center}
%\begin{tabular}{|l|c|}
\begin{tabular}{ll}
\hline
Type & Link\\
\hline\hline
test (1) & \url{https://youtu.be/TKHTHPxFAd4} \\
test (2) & \url{https://youtu.be/6QHP7xhcYx0} \\
test (3) & \url{https://youtu.be/xpUeUa8lZzo} \\
test (4) &\url{https://youtu.be/M_a0DJ5LF5Y} \\
night & \url{https://youtu.be/ezfU1KvDeCA} \\
\hline
\end{tabular}
\end{center}
\caption{Links to videos comparing the DVS-only detector and the RRC (fine-tuned). The threshold for displaying detections is 0.5.}
\label{table:vid}
\end{table}

\begin{figure*}
\begin{center}
   \includegraphics[width=1.0\linewidth]{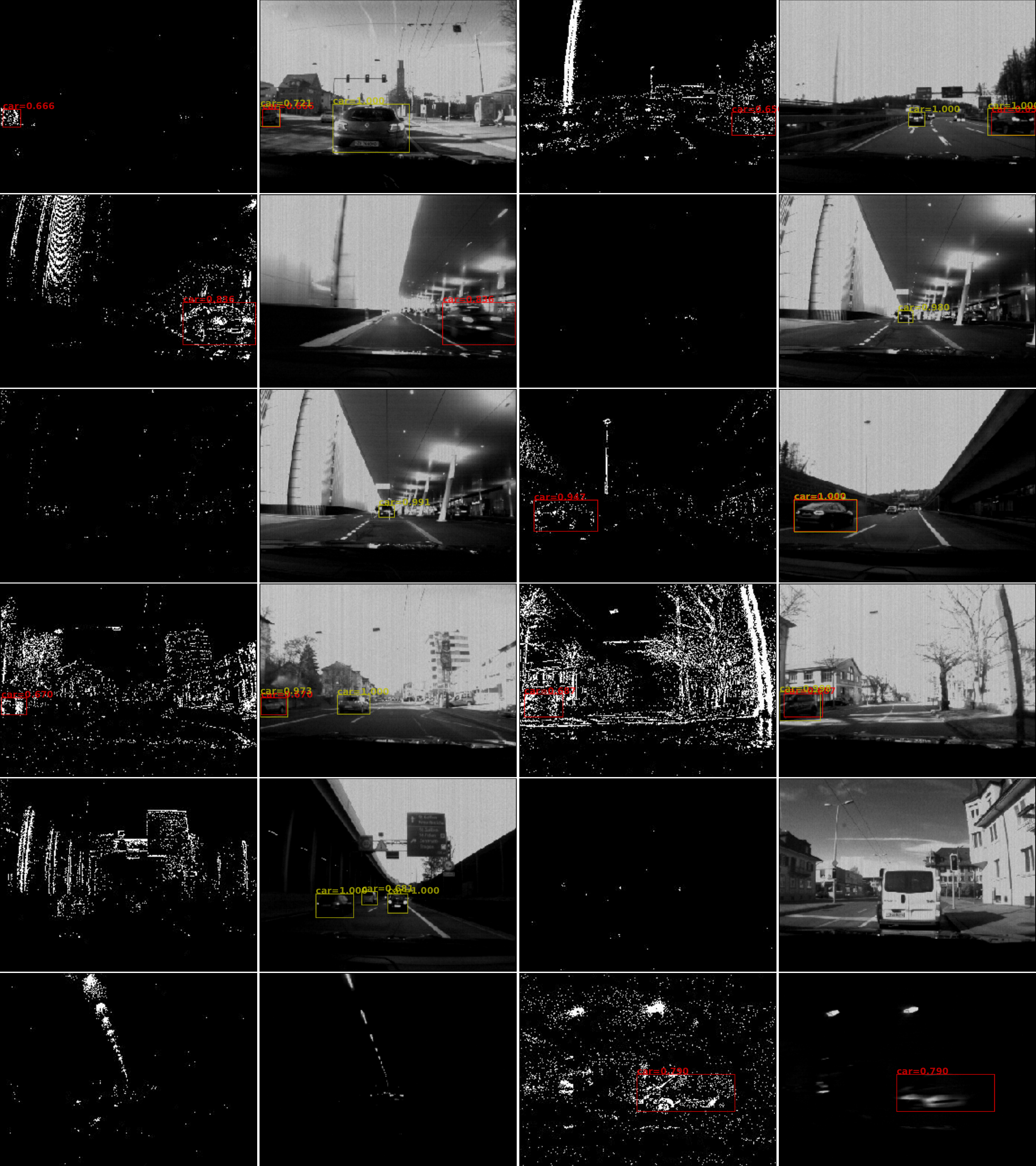}
\end{center}
   \caption{(Best viewed in color and zoomed in) Randomly selected images from day and evening scenes (test set). Images come in pairs: The left image of each pair is the DVS image with bounding boxes (in red) produced by DVS-only detection, while the right image of each pair is the APS image with DVS-only bounding boxes (in red) copied over and the RRC detections (in yellow) for comparison. First row, first pair: DVS fails to detect a stationary car. Second row, first pair: DVS-only detector detects a car despite motion blur, but the RRC fails to do so. Notice that in the DVS image, the edges of the car are still reasonably distinct. Second row, second pair: An example where a car in the far-field does not trigger a response in the DVS. Last row, second pair: Despite dim lighting and motion blur, the edges of the car are still visible on the DVS image and hence the DVS-only detector picks it up, while the RRC misses it.}
\label{fig:dayscene}
\end{figure*}

\begin{figure*}[t]
\begin{center}
   \includegraphics[width=1.0\linewidth]{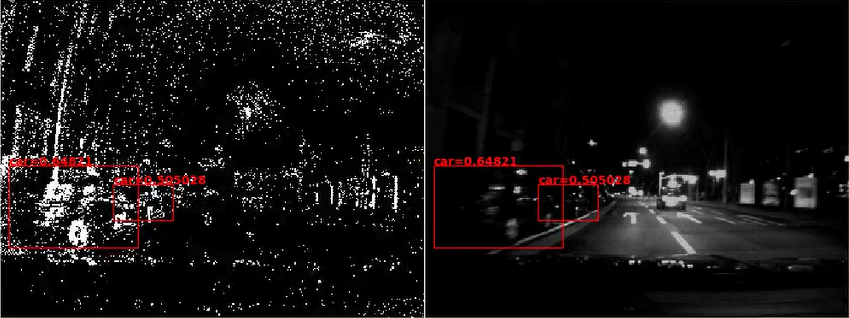}
\end{center}
   \caption{Left: Night scene on the DVS sensor, with bounding boxes produced by the DVS-only detector. Right: APS sensor image, with bounding boxes copied from DVS-only detection for comparison. The RRC did not produce any detections for this image. Notice that the edges of the cars are visible in the DVS image, but barely visible in the APS image.}
\label{fig:nightscene}
\end{figure*}

Though we used the sigmoid representation for training our detector, the following images from the dynamic vision sensor are displayed in the \textit{binary representation} for easier viewing, where each pixel in the frame takes the value

\begin{equation}
b(x) =\begin{cases} 255 &\mbox{if } x \neq 0 \\
0 & \mbox{if } x = 0 \end{cases},
\end{equation}

\noindent where $x$ is the sum of the polarities of the events in the 10 ms interval. The numbers above the bounding boxes indicate the confidence, and the threshold for displaying the bounding boxes on the following images and videos is 0.5. All bounding boxes shown are a result of the fine-tuned RRC and the DVS which is trained on its pseudo-labels. Links to videos can be found in table \ref{table:vid}, and the reader is strongly encouraged to randomly sample clips from all videos to gauge the performance of the DVS-only detector.

\smallskip

\noindent\textbf{Daytime and evening detections}\quad Randomly sampled images from the test sets are shown in Figure~\ref{fig:dayscene}, and these highlight the main sources of errors. While the CNN is able to detect cars in the near-field, cars in the far-field and cars moving at the same velocity as the camera (hence zero relative velocity) only show up on the DVS images as thin outlines at best and as such are not detected by our CNN. This explains why the fraction of objects detected by the DVS is not 100\%.

\smallskip

\noindent\textbf{Overcoming motion blur}\quad In the first pair of row 2 and second pair of row 6 of Figure~\ref{fig:dayscene}, we see the high temporal resolution of the dynamic vision sensor in action. The camera is moving fast and as a result, the features captured by the frame-based camera are blurred, whereas the features captured by the dynamic vision sensor is still reasonably sharp. Our event-based detector managed to detect the cars while the RRC did not produce any detections, reinforcing our motivation for object detection on dynamic vision sensor data. An additional motion blur scene can be found at the 1:30 mark in the video of the third test scene.

\smallskip

\noindent\textbf{Nighttime detections}\quad One key feature of dynamic vision sensors is the high dynamic range which can cope with a wide spectrum of illumination conditions. Figure~\ref{fig:nightscene} shows a night scene (rec1487356509 from the DDD17, at the 2:01:59 mark of the night scene video) where illumination is poor on the left hand side of the lane. The APS sensor barely picks up the cars as they are dark enough to blend into the surrounding, and as such pose a major challenge for conventional frame-based detection. This is confirmed by the fact that the RRC did not manage to detect the cars. However, the DVS can still detect the edges of the cars and as such, the cars on the DVS image are picked out by our DVS-only detector. Considering that the DVS-only detector is trained only on day and evening scenes, the fact that it was able to detect cars at night shows that the detector learnt representations of the cars which are robust to illumination conditions. 

\smallskip

\begin{figure*}[t]
\begin{center}
   \includegraphics[width=1.0\linewidth]{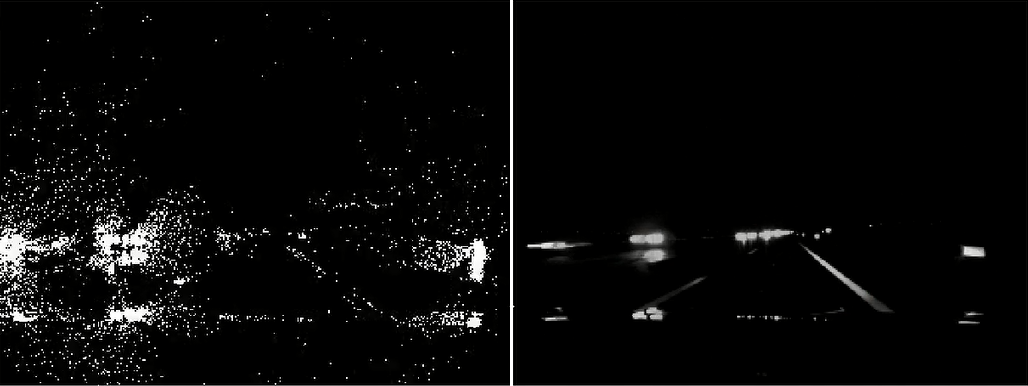}
\end{center}
   \caption{Left: DVS image. Right: APS image. This is a highway scene where we are only able to see the headlights of the car and nothing else. Both the DVS-only detector and the RRC fail to produce detections in such a scenario.}
\label{fig:headlights}
\end{figure*}

\noindent\textbf{Limitations}\quad In Figure~\ref{fig:headlights}, we see an example where our approach fails. This scene is on a highway at night (also from rec1487356509), where the light source is dominated by the headlights of the cars. As the CNN is trained on DVS images of cars in the day and evening scenes, it learns the features that are visible in the day and in the evening (\eg edges of the car) and it does not learn the features of the headlights. To learn such features, we require labeled data which might be hard to obtain from the pseudo-labeling method because conventional CNNs do not work well on images with poor illumination conditions. This strongly suggests that the na$\ddot{\i}$ve approach of binning DVS data and creating images is not sufficient to represent the data.

\section{Discussion}

Our implementation is largely unoptimized, and the average precision can be increased via many ways. For example, we can fine-tune the threshold to keep pseudo-labels for training, the network and learning hyper-parameters of the DVS-only detector and explore other representations of the DVS data (\eg possibly binning the data by a fixed number of events). We can also combine detection results with tracking methods such as particle filter~\cite{Liu2016} or those developed for dynamic vision sensor data~\cite{LagorceTrack, NiTrack,TedaldiTrack}.

In Figure \ref{fig:dayscene}, we saw how our CNN missed detections of cars that are far away, because the pixels that spike are sparsely distributed and possibly drowned out by noise. This issue can be solved via a few ways. For instance, using a higher resolution camera will allow for more pixels to capture the features of the car. However, this approach misses the point of using an dynamic vision sensor--The output of dynamic vision sensors is \textit{intended} to be sparse, because it captures changes in the scene rather than the entire scene itself. The next step is to move away from a frame-based approach when analyzing dynamic vision sensor data, and towards an entirely event-based approach, i.e. use an algorithm that accepts sparse dynamic vision sensor data, and takes temporal information into account. For example, we can combine the event-based ROI approach in~\cite{Liu2016} with event-based recognition approaches such as HOTS~\cite{HOTS} or spiking neural networks~\cite{SNN2,SNN3,SNN1}. These event-based recognition approaches can also be trained with pseudo-labels.

\section{Conclusions and Future Work}

In all, we have presented two main contributions. First, we showed for the first time high speed (100 FPS) detection of a realistic object (car) in a real scenario with various backgrounds and distracting objects due to camera ego-motion, purely from dynamic vision sensor data. Previous work on event-based detection/recognition have only focused on recognizing simple objects such as numbers, or detecting objects in the absence of ego-motion, and the most realistic work is on detecting a robot in a controlled lab environment~\cite{Liu2016}. Our technique showed reasonable success with detections in day and night scenes, however it failed to detect cars when the headlights are bright enough to distort the features, or when the cars are too far away and show up as very sparse pixels. We suggested approaches to overcome these problems, such as using a fully event-based framework. Second, we showed that our trained CNN can detect cars despite motion blur and poor lighting without explicit training on such scenes, and even cars which were not detected by the RRC in ordinary conditions--This proves that the CNN learnt robust representations of cars from pseudo-labels.

Future work includes implementing spiking neural networks on neuromorphic computing hardware, which could potentially bring a 70 times increase in power efficiency compared to traditional hardware~\cite{osswald2017}. We see value in performing event-based image segmentation because it could boost detection performance and overcome the headlights problem in Figure~\ref{fig:headlights} (\eg if we detect an object on the road, then the object is more likely to be a car even though the DVS detector only sees headlights).

We hope that this work will encourage researchers to use pseudo-labels for supervised learning techniques on dynamic vision sensor data and advance the frontiers of this field, and to publish more data sets containing synchronized DVS and APS modalities.

{\small
\bibliographystyle{ieee}
\bibliography{egbib}
}

\end{document}